\pdfoutput=1

\documentclass[11pt]{article}

\usepackage{acl}

\usepackage{times}
\usepackage{latexsym}

\usepackage[T1]{fontenc}

\usepackage[utf8]{inputenc}

\usepackage{microtype}

\usepackage{booktabs}
\usepackage{enumitem}
\usepackage{color}
\usepackage{tikz-dependency}
\usepackage{amsfonts}
\usepackage{amsmath}
\usepackage{bbold}
\usepackage{inconsolata}
\usepackage{multirow}
\usepackage{makecell}
\usepackage{pifont}

\newcommand{\coconut}{\includegraphics[width=0.02\textwidth]{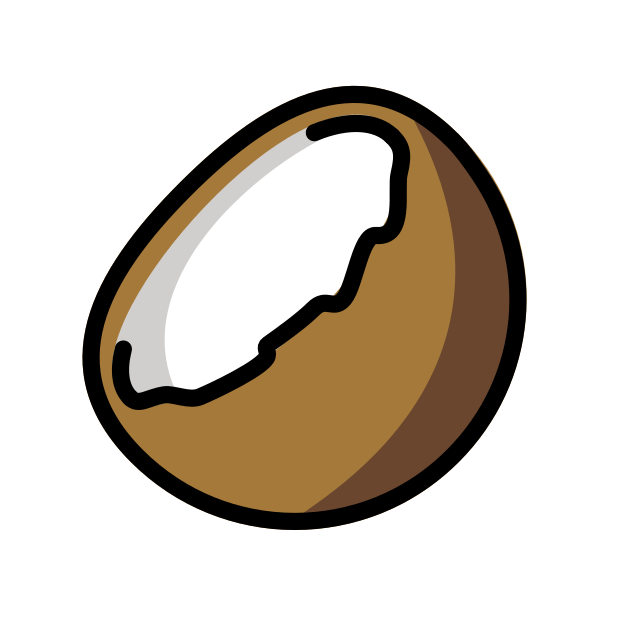}}
\newcommand{\kiwi}{\includegraphics[width=0.02\textwidth]{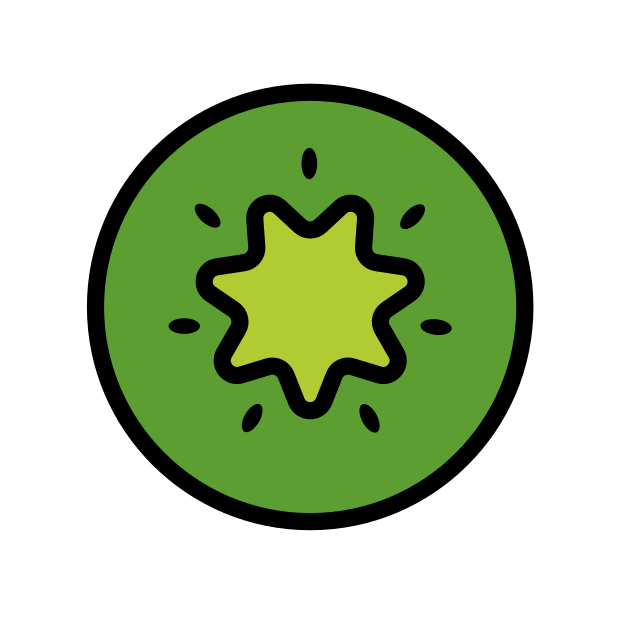}}
\newcommand{\lemon}{\includegraphics[width=0.02\textwidth]{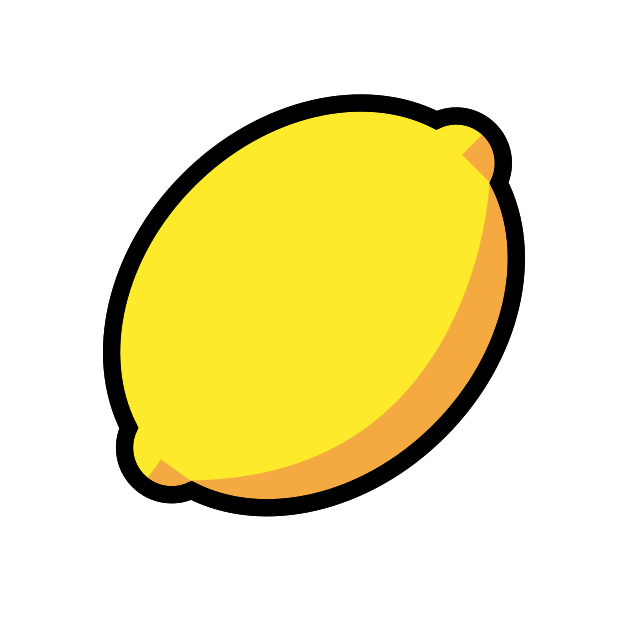}}

\title{Structural Bias for Aspect Sentiment Triplet Extraction}

\author{
  Chen Zhang\textsuperscript{\coconut}, Lei Ren\textsuperscript{\kiwi}, Fang Ma\textsuperscript{\coconut}, Jingang Wang\textsuperscript{\kiwi\lemon}, Wei Wu\textsuperscript{\kiwi}, Dawei Song\textsuperscript{\coconut\lemon}\Thanks{\textsuperscript{\lemon}Jingang Wang and Dawei Song are the corresponding authors.} \\
  \textsuperscript{\coconut}Beijing Institute of Technology \\
  \texttt{\{czhang,mfang,dwsong\}@bit.edu.cn} \\
  \textsuperscript{\kiwi}Meituan NLP \\
  \texttt{\{wangjingang02,wuwei30\}@meituan.com} \\
  \texttt{renlei\_work@163.com} \\
}

\makeatletter
\def\thanks#1{\protected@xdef\@thanks{\@thanks
    \protect\footnotetext{#1}}}
\makeatother

\begin{document}

\maketitle

\begin{abstract}
Structural bias has recently been exploited for aspect sentiment triplet extraction (ASTE) and led to improved performance. On the other hand, it is recognized that explicitly incorporating structural bias would have a negative impact on efficiency, whereas pretrained language models (PLMs) can already capture implicit structures. Thus, a natural question arises: Is structural bias still a necessity in the context of PLMs? To answer the question, we propose to address the efficiency issues by using an adapter to integrate structural bias in the PLM and using a cheap-to-compute relative position structure in place of the syntactic dependency structure. Benchmarking evaluation is conducted on the SemEval datasets. The results show that our proposed structural adapter is beneficial to PLMs and achieves state-of-the-art performance over a range of strong baselines, yet with a light parameter demand and low latency. Meanwhile, we give rise to the concern that the current evaluation default with data of small scale is under-confident. Consequently, we release a large-scale dataset for ASTE. The results on the new dataset hint that the structural adapter is confidently effective and efficient to a large scale. Overall, we draw the conclusion that structural bias shall still be a necessity even with PLMs.\footnote{Code and data are available at \url{https://github.com/GeneZC/StructBias}.}
\end{abstract}

\section{Introduction}

\begin{figure}[t]
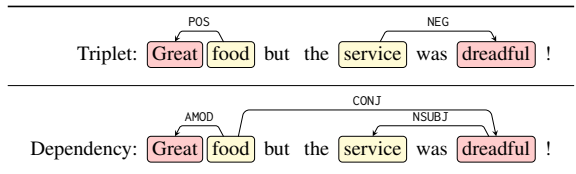

    \centering
    \resizebox{.47\textwidth}{!}{
    \begin{tabular}{rc}
    \toprule
    \begin{dependency}[text only label, label style={above}]
    \begin{deptext}[column sep=.1cm]
    Triplet: \& Great \& food \& but \& the \& service \& was \& dreadful \& ! \\ 
    \end{deptext} 
    \depedge[edge height=.2cm]{3}{2}{\texttt{POS}} \depedge[edge height=.2cm]{6}{8}{\texttt{NEG}}
    \wordgroup[group style={fill=red!20}]{1}{2}{2}{op1}
    \wordgroup[group style={fill=yellow!20}]{1}{3}{3}{ap1}
    \wordgroup[group style={fill=red!20}]{1}{8}{8}{op1}
    \wordgroup[group style={fill=yellow!20}]{1}{6}{6}{ap1}
    \end{dependency} \\
    \midrule
    \begin{dependency}[text only label, label style={above}]
    \begin{deptext}[column sep=.1cm]
    Dependency: \& Great \& food \& but \& the \& service \& was \& dreadful \& ! \\ 
    \end{deptext} 
    \depedge[edge height=.2cm]{3}{2}{\texttt{AMOD}} \depedge[edge height=.2cm]{8}{6}{\texttt{NSUBJ}}
    \depedge[edge height=.5cm]{3}{8}{\texttt{CONJ}}
    \wordgroup[group style={fill=red!20}]{1}{2}{2}{op1}
    \wordgroup[group style={fill=yellow!20}]{1}{3}{3}{ap1}
    \wordgroup[group style={fill=red!20}]{1}{8}{8}{op1}
    \wordgroup[group style={fill=yellow!20}]{1}{6}{6}{ap1}
    \end{dependency} \\
    \bottomrule
    \end{tabular}}
    \caption{Triplet (top) and dependency structures (bottom) of an illustrative sentence. Spans shaded in yellow are aspects, spans shaded in red are opinions, and arcs indicate either sentiment or structural relations. Irrelevant structural relations are neglected for brevity.}
    \label{fig1}
\end{figure}

Aspect sentiment triplet extraction (ASTE) is a task central to fine-grained opinion mining. Compared to aspect sentiment classification that only aims to predict sentiment polarities for various aspects, ASTE instead extracts descriptive opinion units in the form of triplets (i.e., aspect-opinion-sentiment tuples). For example, (food, great, \texttt{POS}) and (service, dreadful, \texttt{NEG}) are aspect sentiment triplets for the sentence in Figure~\ref{fig1} (top), where \{\texttt{POS}, \texttt{NEG}, \texttt{NEU}\} respectively represent \{positive, negative, neutral\}.

While ASTE can be generally tackled with neural models in either a pipeline manner~\citep{Peng20} or a multi-task manner~\citep{Xu20,Zhang20,Wu20,Chen21a,Xu21}, the aspect sentiment triplets can be rather derivable from dependency structures (e.g., syntactic dependency trees) with hand-crafted rules~\citep{Wu09,Sun17}. For the example in Figure~\ref{fig1} (bottom), the triplets can be recognized via certain structural dependency relations.\footnote{Please see \url{https://downloads.cs.stanford.edu/nlp/software/dependencies_manual.pdf} for what these structural relations exactly stand for.}  Various studies are motivated by this intuition and exploit dependency bias to enhance neural ASTE models~\citep{Chen21b}, yet without a necessary comparison with pretrained language models (PLMs). On the other hand, recent advances find that using PLMs can already achieve compelling performance~\citep{Yan21,Zhang21,Huang21} owing to implicit structures captured by PLMs~\citep{Wu20b}. It signals that, compared with PLMs, explicit structural biases such as dependency bias, may become cumbersome~\citep{Dai21} due to \textit{parameter inefficiency} and \textit{latency inefficiency}. That is, combining dependencies into models can require redundant parameters to achieve structure encoding, while producing structures can also require increased latency to achieve external parsing. Therefore, a critical question naturally arises: Is structural bias still \textit{a necessity} for ASTE in the context of PLMs?

In this paper, we aspire to answer the question from two perspectives: 1) whether structural bias can be incorporated into PLMs in a flexible way in terms of both parameter and latency efficiency; and 2) whether structural bias can enhance PLMs for ASTE.

To boost the parameter efficiency, we develop the idea of \textit{adapter} and put forward a parameter-efficient adapter that can incorporate structural bias. The adapter~\citep{Houlsby19} was proposed initially to integrate additional modules into PLMs and enable PLMs to leverage inductive bias efficiently. Although feasible, such adapters can be far from lightweight. For example, \citet{Liu21} introduces a series of linear transformations in their proposed adapter which involve numerous parameters. In contrast, instead of introducing carefully-designed plugins, we propose to use structured attention maps induced with structures, to additively impact the raw attention maps in self-attention, thus requiring only a tiny amount of incremental parameters.

To improve the latency efficiency, we argue that dependency distance is a sufficient simplification of the dependency graph since the simplification has been proven equally powerful in downstream tasks like aspect sentiment classification~\citep{Zhang19}. On this basis, we further propose to use \textit{relative distance} as an alternative to the dependency distance. The intuition lies in the observation that opinions predominantly locate closely to their corresponding aspects~\citep{Xu20, Ma21}, and thus we posit that using relative distance bias would suffice for the purpose of ASTE. In fact, the relative distance is also exhibited to bring merits to the transformer architectures in previous work~\citep{Shaw18,Raffel20}. As the relative distance can be obtained with cheap operations in lower latency, the latency efficiency issue is thereby resolved.

We conduct a benchmarking comparative study on the SemEval datasets~\citep{Pontiki14}. The results show that models with the proposed structural adapter achieve the state-of-the-art (SOTA) performance compared with an array of strong baselines, indicating that incorporating structural bias is beneficial to PLMs. We also conduct a further study on how the relative distance-derived structural adapter overwhelms its alternatives. The results demonstrate that the structural adapter is an appealing choice. Specifically, our structural adapter realizes a 1,000$\times$ scale-down in terms of incremental parameters and a 1,000$\times$ speed-up of distance derivation.

In summary, structural bias can be flexibly incorporated into PLMs and improve both parameter and latency efficiency. The structural adapter is imposed with only a light parameter demand. The relative distance can be implemented with low latency. Moreover, the structural adapter vastly improves the SOTA performance. Therefore, structural bias is still be a necessity even in the context of PLMs to achieve a better ASTE performance. 

Last but not least, in the view that current benchmarks are of small scales, we create a large-scale ASTE dataset termed \textbb{Lasted}. \textbb{Lasted} is collected from one of the largest review platform in China, namely DianPing.\footnote{Please see \url{https://www.dianping.com/}.} The dataset will be released to facilitate a more confident evaluation for ASTE and other possible research directions. The results on \textbb{Lasted} hint that structural adapter confidently improves the performance. Furthermore, compared with the results on the SemEval datasets, the model performance generally tends to be lower on this dataset, suggesting that the large-scale deployment of ASTE systems is still challenging.

\section{Methodology}

\subsection{Task Formulation}
\label{sec2.1}

Given a sequence of tokens $\{t_i\}_{i=1}^{n}$ as input, ASTE requires a model to output a set of triplets $\{(a,o,s)_i\}_{i=1}^{m}$, where $a$, $o$, $s$ are the aspect, opinion, and sentiment, respectively. Concretely, an aspect $a$ can be decomposed to two elements, i.e. $(a_0, a_1)$, that separately denote the start and end positions. Likewise, an opinion $o$ can be decomposed similarly.

\subsection{PLM with Structural Adapter}

When a PLM is employed as the backbone, the tokens are first transformed to embeddings, and then manipulated by subsequent transformer blocks. While the PLM can capture semantic interactions, we additionally present how to include structural interactions with a structural adapter.

\subsubsection{Embedding}

The tokens are generally augmented and encoded with the PLM. For example, if the PLM being used is a \texttt{BERT}~\citep{Devlin19}, the tokens should be augmented as:
\begin{equation}\nonumber
    \mathsf{[CLS]} \quad t_1 \quad \dots \quad t_i \quad \dots \quad t_n \quad \mathsf{[SEP]}
\end{equation}
After that, the augmented tokens are converted to embeddings $\{\mathbf{t}_i\}_{i=0}^{n+1}$.

\subsubsection{Transformer Block}

The input embeddings are operated by the succeeding transformer blocks~\citep{Vaswani17}, each of which consists of a self-attention module and a feed-forward network module. The self-attention module is typically organized in a query-key-value formulation. Specifically, for any input $\{\mathbf{x}_i\}_{i=1}^{n}$, the output can be roughly written as:
\begin{equation}
    \begin{aligned}
    \mathbf{z}_i=\sum_{j=1}^{n}&\alpha_{ij}(\mathbf{x}_j\mathbf{W}_{V}) \quad \alpha_{ij}=\text{softmax}_{j}(e_{ij}) \\
    &e_{ij}=\frac{\mathbf{x}_{i}\mathbf{W}_{Q}(\mathbf{x}_{j}\mathbf{W}_{K})^{\top}}{\sqrt{d}}
    \end{aligned}
\end{equation}
Here, we omit special tokens and multiple heads for simplicity. The parameters $\mathbf{W}_{Q}$, $\mathbf{W}_{K}$ and $\mathbf{W}_{V}$ are learnable linear transformations for the query, key, and value. $d$ is the head dimensionality.

\subsubsection{Structural Adapter}

\begin{figure}[t]
    \centering
    \includegraphics[width=0.37\textwidth]{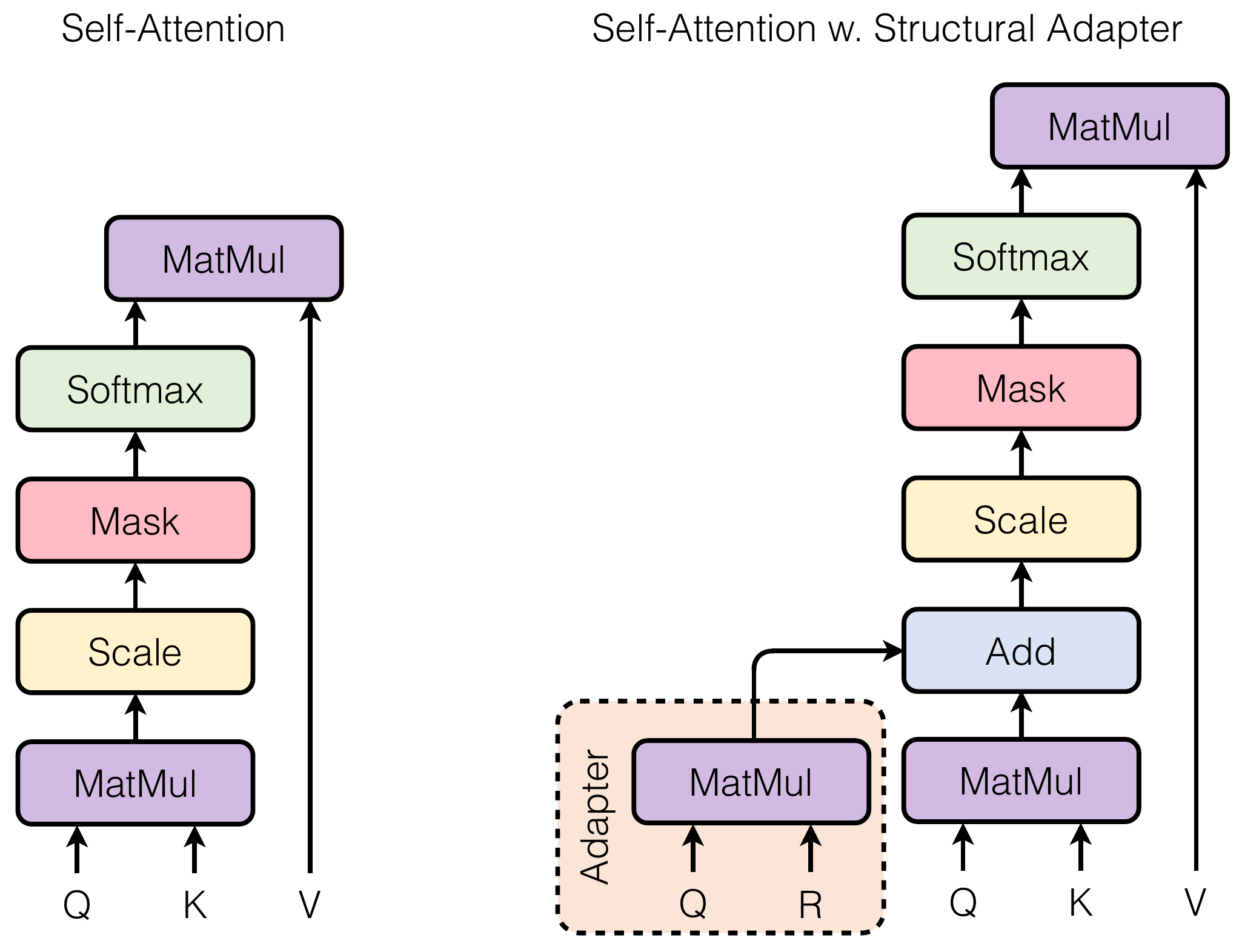}
    \caption{Differences between self-attention (left) and self-attention with the structural adapter applied (right). \textsf{\small Q}, \textsf{\small K}, and \textsf{\small V} separately stand for linear-transformed queries, keys, and values. \textsf{\small R} stands for distances.}
    \label{fig2}
\end{figure}

In order to integrate the dependency or relative distance into the self-attention, the structural adapter is imposed to derive structured attention maps to bias the raw attention maps induced with the self-attention additively. The procedure is depicted as below:
\begin{equation}
    \begin{aligned}
    &e_{ij}=\frac{\mathbf{x}_{i}\mathbf{W}_{Q}(\mathbf{x}_{j}\mathbf{W}_{K}+\mathbb{r}_{ij})^{\top}}{\sqrt{d}} \\
    &=\underbrace{\frac{\mathbf{x}_{i}\mathbf{W}_{Q}(\mathbf{x}_{j}\mathbf{W}_{K})^{\top}}{\sqrt{d}}}_{\text{raw attention map}}+\underbrace{\frac{\mathbf{x}_{i}\mathbf{W}_{Q}\mathbb{r}_{ij}^{\top}}{\sqrt{d}}}_{\text{structured attention map}} 
    \end{aligned}
\end{equation}
where $\mathbb{r}_{ij}$ indicates the distance embedding between two tokens $t_i$ and $t_j$. It is also noteworthy that each relation embedding is shared across different heads, but kept independent from one layer to another layer. This behavior is inspired by~\citet{Shaw18}, which is originally proposed to encode the relative positions but found to be applicable to encode arbitrary relations~\citep{Wang20}. The differences between self-attention and self-attention with the structural adapter applied are shown in Figure~\ref{fig2}.

\subsubsection{Distance Derivation}

\begin{figure}[t]
    \centering
    \includegraphics[width=0.47\textwidth]{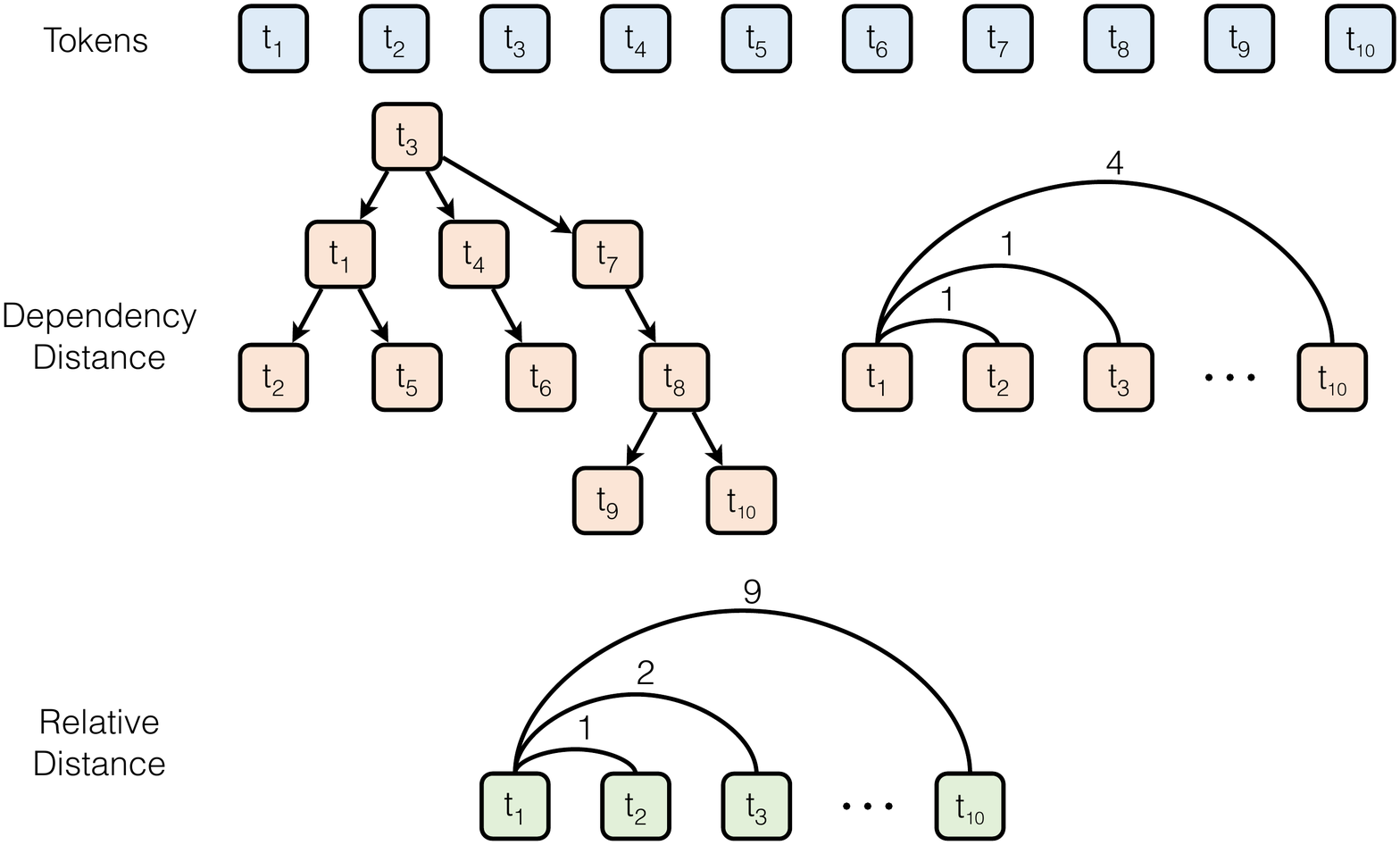}
    \caption{Derivation procedures of dependency distance (middle) and relative distance (bottom) given a sequence of tokens (top). Large distance values like \textsf{\small 9} in this example may be clipped by the threshold $\tau$. And distances will be made directional when employed, e.g., distance from \textsf{\small t\textsubscript{1}} to \textsf{\small t\textsubscript{2}} is set to \textsf{\small 1} while that from \textsf{\small t\textsubscript{2}} to \textsf{\small t\textsubscript{1}} is actually set to \textsf{\small -1}.}
    \label{fig3}
\end{figure}

Specifically, the dependency distance between two tokens is obtained by computing the shortest distance on the dependency graph with the \textsf{\small networkx} toolkit,\footnote{Please see \url{https://networkx.org/} for more information.} and the dependency graph is produced with an off-the-shelf dependency parser \textsf{\small stanza}~\citep{Qi20}.\footnote{Please see \url{https://stanfordnlp.github.io/stanza/} for more information.} The relative distance between two tokens can be yielded by enumerating the number of tokens lying in-between. 

We follow the de facto implementation that treats the distance from $t_i$ to $t_j$ different from that from $t_j$ to $t_i$. We assign one to positive and the other to negative~\citep{Raffel20}. We also manually set a distance threshold that denotes the maximum distance $\tau$. In doing so, we intend to avoid introducing too many parameters while maintaining as much information as possible. Henceforth, we will refer to the structural adapter with the dependency distance and relative distance as \textsc{\small StructApt-Dep} and \textsc{\small StructApt-Rel}. The derivation of both dependency distance and relative distance is illustrated in Figure~\ref{fig3}.

\subsection{Triplet Parser}

Learning from two multi-task triplet parsing architectures MTL~\citep{Zhang20} and GTS~\citep{Wu20}, we establish a triplet parser that comprises two independent taggers (i.e., one for aspect and the other for opinion tagging), a sentiment scorer, and a triplet decoder. Conceptually, the two taggers are used to uncover continuous tokens that form an aspect or opinion span. The sentiment scorer is used to determine the token-level sentiment relation (if there is one) between two candidate tokens. Moreover, the triplet decoder produces triplets by gathering the information from the taggers and the sentiment scorer. 

\subsubsection{Aspect and Opinion Taggers}

Following MTL, the taggers generate aspect and opinion tags in \{\texttt{B,I,O}\} format, after which the aspect and opinion spans are inferred with \{\texttt{B,I,O}\} tags. Presuming the hidden states of the PLM are $\{\mathbf{h}_i\}_{i=1}^n$ in spite of augmented tokens, the taggers are depicted as:
\begin{equation}
    \begin{aligned}
    \mathbf{r}_i^{(a)}=\text{ReLU}(\mathbf{W}_1^{(a)}\mathbf{h}_i+\mathbf{b}_1^{(a)}) \\
    \mathbf{y}_i^{(a)}=\text{softmax}(\mathbf{W}_2^{(a)}\mathbf{r}_i^{(a)}+\mathbf{b}_2^{(a)}) \\
    \mathbf{r}_i^{(o)}=\text{ReLU}(\mathbf{W}_1^{(o)}\mathbf{h}_i+\mathbf{b}_1^{(o)}) \\
    \mathbf{y}_i^{(o)}=\text{softmax}(\mathbf{W}_2^{(o)}\mathbf{r}_i^{(o)}+\mathbf{b}_2^{(o)}) \\
    \end{aligned}
\end{equation}
where \{$\mathbf{W}_1^{(a)},\mathbf{b}_1^{(a)},\mathbf{W}_2^{(a)},\mathbf{b}_2^{(a)}$\}, \{$\mathbf{W}_1^{(o)},\mathbf{b}_1^{(o)},\mathbf{W}_2^{(o)},\mathbf{b}_2^{(o)}$\} are two sets of weights and biases for two feed-forward networks customized to aspect and opinion tagging.

\subsubsection{Sentiment Scorer}

The sentiment scorer produces token-level sentiment relations among all tokens. In addition to \{\texttt{POS}, \texttt{NEG},  \texttt{NEU}\}, there is also a \texttt{NONE} relation to account for the case of no relation. Unlike the sentiment scorer in MTL and GTS that only predicts uni-directional sentiment relations, we present a sentiment scorer that predicts bi-directional sentiment relations. The uni-directional relation means: a sentiment relation between an aspect token and an opinion token is always directed from the aspect token to the opinion token. In contrast, the bi-directional means: a sentiment relation is both directed from the aspect token to the opinion token and directed from the opinion token to the aspect token. This behavior allows more information to be transduced to the subsequent triplet decoding process to alleviate potential errors. Similarly, the sentiment scorer can be described as:
\begin{equation}
    \begin{aligned}
    \mathbf{r}_i^{(h)}=\text{ReLU}(\mathbf{W}_1^{(h)}\mathbf{h}_i+\mathbf{b}_1^{(h)}) \\
    \mathbf{r}_i^{(d)}=\text{ReLU}(\mathbf{W}_1^{(d)}\mathbf{h}_i+\mathbf{b}_1^{(d)}) \\
    \mathbf{y}_{i,j}^{(s)}=\text{softmax}(\mathbf{r}_i^{(h)\top}\mathbf{W}_2^{(s)}\mathbf{r}_j^{(d)}+ \\
    {\mathbf{W}_2^{(h)}}\mathbf{r}_i^{(h)}+{\mathbf{W}_2^{(d)}}\mathbf{r}_j^{(d)}+\mathbf{b}_2^{(s)})
    \end{aligned}
\end{equation}
Here, \{$\mathbf{W}_1^{(h)},\mathbf{b}_1^{(h)}$\},\{$\mathbf{W}_1^{(d)},\mathbf{b}_1^{(d)}$\} are weights and biases separately for two feed-forward networks yielding head and dependent representations. These head and dependent representations are then organized in a biaffine manner~\citep{Dozat17}, where \{$\mathbf{W}_2^{(h)},\mathbf{b}_2^{(h)},\mathbf{W}_2^{(d)},\mathbf{b}_2^{(d)}$\} are weights and biases. The biaffine module predicts both aspect-to-opinion and opinion-to-aspect relations at the same time, since either aspects or opinions can be heads or dependents interchangeably.

\begin{figure}[t]
    \centering
    \includegraphics[width=0.37\textwidth]{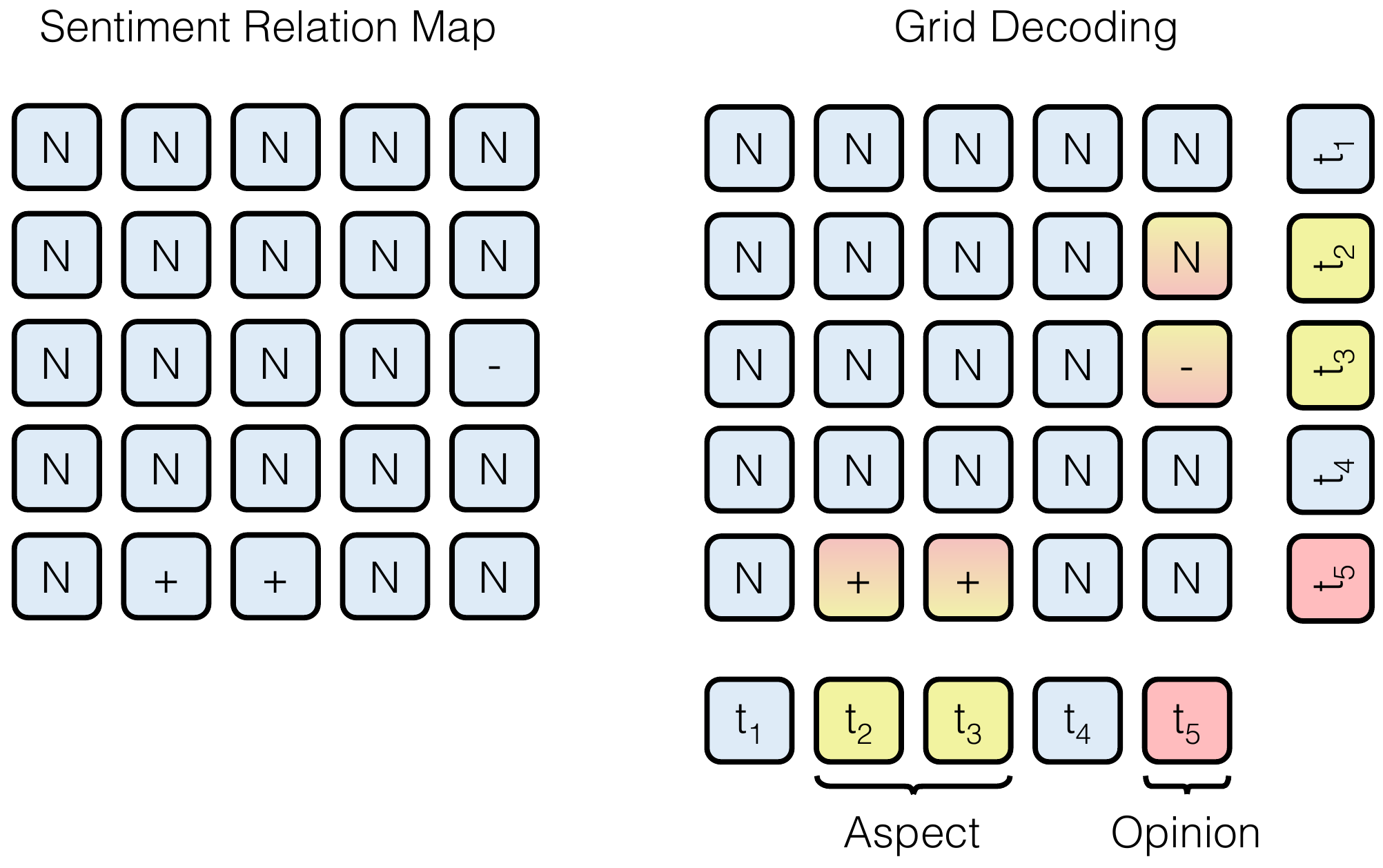}
    \caption{Sentiment relation map (left) and grid decoding algorithm (right). \textsf{\small N}, \textsf{\small +}, \textsf{\small -} respectively are short for \texttt{NONE}, \texttt{POS}, \texttt{NEG} relations. Gradient colors indicate either aspect-to-opinion relations or opinion-to-aspect relations within the bi-directional interplay. The grid decoding algorithm refers to tagged aspects and opinions to index the sentiment relation map.}
    \label{fig4}
\end{figure}

Additionally, $\mathbf{y}^{(s)}$ refers to a sentiment probability map where $\mathbf{y}_{i,j}^{(s)}$ indicates a probability over 4 sentiment relations from the $i$-th token to the $j$-th one. If we apply the argmax operation on the probability map, we get a sentiment relation map. An example of the sentiment relation map is given in Figure~\ref{fig4} (left).

\subsubsection{Triplet Decoder}

Span-level sentiment relations are viable by searching for the most frequent sentiment relation in the set of indexed sentiment relations. Assume there are 2 tokens in a predicted target span and 1 token in a predicted opinion span, then we say there are 2*(2*1)=4 indexed sentiment relations between the two spans with the bi-directional interplay. Finding the most frequent sentiment relation in inclusive sentiment relations produced with the sentiment relation map gives the sentiment relation between the two spans. The algorithm is detailed in Figure~\ref{fig4} (right). As the bi-directional interplay is considered, the potential error (i.e., producing a \texttt{NONE} or \texttt{NEG} relation instead of a \texttt{POS} one.) in the example is alleviated.

Since the whole multi-task learning framework is generally borrowed from MTL while the triplet decoding strategy is adapted from the grid decoding in GTS, we name the proposed model as \underline{Mu}lti-task learning with \underline{G}rid decoding (MuG).

\subsection{Finetuning}

The PLM, the structural adapter, and the triplet parser can be jointly optimized by minimizing an overall objective that contains two sources of losses, i.e., tagging loss $\mathcal{L}_\text{t}$ and parsing loss $\mathcal{L}_\text{p}$. Both losses can be measured by the cross-entropy functions. The joint objective is formulated as follows:
\begin{equation}
    \min_{\theta}\mathcal{L}=\min_{\theta}\mathcal{L}_\text{t}+\mathcal{L}_\text{p}
\end{equation}
where $\theta$ stands for all parameters, which might be disassembled to \{$\theta_{\text{PLM}},\theta_{\text{Adapter}}, \theta_{\text{Parser}}$\}.

\section{Benchmarking Evaluation}

\subsection{Data}

We conduct a comparative study on 4 benchmarking datasets from \textbf{SemEval} 2014, 2015, and 2016~\citep{Pontiki14}, in which one contains data from laptop domain (\textbf{L}) and the other three contain data from restaurant domain (\textbf{R}). The triplet annotations are obtained from~\citet{Xu20}. The statistics of these datasets are displayed in Appendix~\ref{app1}.

\subsection{Models}

We compare a wide range of baseline models with varying backbones (e.g., \texttt{BiLSTM}, \texttt{BERT}) and different paradigms (i.e., extractive vs. generative). We list these baselines according to their paradigms as below:

\begin{description}
    \item[Extractive Paradigm]\quad
    \begin{itemize}[leftmargin=0pt]
        \item KWHW~\citep{Peng20} is a pipeline system that first extracts aspect-sentiment pairs and opinions, and then pairs them in a binary manner.
        \item JET$^o$~\citep{Xu20} is a position-aware sequence tagging system that jointly extracts triplets.
        \item MTL~\citep{Zhang20} is a multi-task learning system which realizes aspect and opinion extraction with tagging while sentiment relation extraction with parsing.
        \item GTS~\citep{Wu20} transforms the triplet extraction problem as a grid tagging problem and achieves the extraction via a grid decoding algorithm.
        \item Span~\citep{Xu21} is a span-level triplet extraction system that learns span-level interactions for a more accurate triplet prediction.
    \end{itemize}
    \item[Generative Paradigm]\quad
    \begin{itemize}[leftmargin=0pt]
        \item UGF~\citep{Yan21} is a unified generative system based on \texttt{BART}~\citep{Lewis20} for all sub-tasks in aspect sentiment analysis.
        \item GAS~\citep{Zhang21} likewise is built upon \texttt{T5}~\citep{Raffel20} where extractive constraints are applied to the decoding space. 
    \end{itemize}
\end{description}

On another note, we have some variants in the comparative study to facilitate the understanding of our adapter. To examine the broad applicability of our structural adapter, we additionally test the structural adapter through the lens of the SOTA extractive systems, namely GTS and Span. To conduct a fair comparison, we initiate the PLM in our model not only with \texttt{BERT}~\citep{Devlin19} but also with \texttt{RoBERTa}~\citep{Liu19} to see whether our model with the adapter is competitive with those enhanced by advanced generative PLMs.

\subsection{Implementation and Metrics}
\label{sec3.3}

Typically, the adapter-based finetuning only tunes $\theta_{\text{Adapter}}$ and $\theta_{\text{Parser}}$, and freezes $\theta_{\text{PLM}}$. As a randomly initialized adapter can be an unsteady factor to the PLM, standard finetuning (i.e., tuning all parameters) can result in performance with high variance~\citep{Houlsby19}. However, sub-optimal phenomenon has been observed in the literature~\citep{Liu21} that such adapter-based finetuning is less promising than standard fine-tuning if the adapter is intended to integrate discrete information (e.g., structural information in our case). Thus, we adapt $\theta_{\text{Adapter}}$ and $\theta_{\text{PLM}}$ to the concerned task via finetuning all parameters (i.e., $\theta_{\text{PLM}}$, $\theta_{\text{Adapter}}$,  and $\theta_{\text{Parser}}$). Other implementation details are listed in Appendix~\ref{app2}.

\begin{table*}[t]
    \centering
    \resizebox{\textwidth}{!}{
    \begin{tabular}{lcccccccccccc}
         \toprule
         \multirow{2}{*}{\textbf{Model}} & \multicolumn{3}{c}{\textbf{L14}} & \multicolumn{3}{c}{\textbf{R14}} & \multicolumn{3}{c}{\textbf{R15}} & \multicolumn{3}{c}{\textbf{R16}} \\
         \cmidrule{2-13}
         & \textbf{P} & \textbf{R} & \textbf{F}\textsubscript{1} & \textbf{P} & \textbf{R} & \textbf{F}\textsubscript{1} & \textbf{P} & \textbf{R} & \textbf{F}\textsubscript{1} & \textbf{P} & \textbf{R} & \textbf{F}\textsubscript{1} \\
         \midrule
         KWHW \texttt{\small BiLSTM}\textsuperscript{*} & 37.38 & 50.38 & 42.87 & 43.24 & 63.66 & 51.46 & 48.07 & 57.51 & 52.32 & 46.96 & 64.24 & 54.21 \\
         JET$^{o}$ \texttt{\small BiLSTM}\textsuperscript{*} & 53.03 & 33.89 & 41.35 & 61.50 & 55.13 & 58.14 & 64.37 & 44.33 & 52.50 & 70.94 & 57.00 & \textbf{63.21} \\
         MTL \texttt{\small BiLSTM}\textsuperscript{\ddag} & 51.00 & 40.07 & 44.81 & 63.87 & 54.76 & 58.90 & 57.50 & 42.56 & 48.73 & 59.03 & 54.84 & 56.73 \\
         GTS \texttt{\small BiLSTM}\textsuperscript{\ddag} & 60.32 & 38.98 & \textbf{47.25} & 71.08 & 56.38 & \textbf{62.85} & 66.60 & 46.91 & \textbf{55.02} & 68.75 & 56.02 & 61.71 \\
         \midrule
         JET$^{o}$ \texttt{\small BERT}\textsuperscript{*} & 55.39 & 47.33 & 51.04 & 70.56 & 55.94 & 62.40 & 64.45 & 51.96 & 57.53 & 70.42 & 58.37 & 63.83 \\
         GTS \texttt{\small BERT}\textsuperscript{\ddag} & 57.09 & 50.33 & 53.48 & 69.49 & 67.75 & 68.59 & 61.59 & 58.21 & 59.81 & 65.75 & 68.32 & 66.99 \\
         \quad w/ \textsc{\small StructApt-Rel} & 57.89 & 51.57 & \underline{54.47} & 68.94 & 68.26 & \underline{68.60} & 62.17 & 58.63 & \underline{60.28} & 66.17 & 69.79 & \underline{67.91} \\
         Span \texttt{\small BERT}\textsuperscript{\ddag} & 62.57 & 56.02 & 59.08 & 71.77 & 70.42 & 71.06 & 62.06 & 63.26 & 62.63 & 68.57 & 71.12 & 69.79 \\
         \quad w/ \textsc{\small StructApt-Rel} & 64.72 & 56.80 & \underline{\textbf{60.47}} & 72.53 & 71.75 & \underline{\textbf{72.13}} & 62.80 & 63.79 & \underline{\textbf{63.17}} & 68.94 & 70.74 & \underline{\textbf{69.80}} \\
         MuG \texttt{\small BERT} & 58.30 & 52.21 & 55.06 & 68.40 & 67.64 & 68.00 & 60.65 & 54.12 & 57.10 & 66.26 & 67.39 & 66.74 \\
         \quad w/ \textsc{\small StructApt-Dep} & 59.39 & 52.95 & \underline{55.95} & 67.69 & 68.90 & \underline{68.27} & 60.74 & 55.77 & \underline{58.11} & 64.73 & 68.33 & 66.45 \\
         \quad w/ \textsc{\small StructApt-Rel} & 59.54 & 52.56 & \underline{55.75} & 68.92 & 68.12 & \underline{68.50} & 59.83 & 56.78 & \underline{58.17} & 65.31 & 68.83 & \underline{67.01} \\
         \midrule
         UGF \texttt{\small BART}\textsuperscript{\dag} & 61.41 & 56.19 & 58.69 & 65.52 & 64.99 & 65.25 & 59.14 & 59.38 & 59.26 & 66.60 & 68.68 & 67.62 \\
         GAS \texttt{\small T5}\textsuperscript{\dag} & -- & -- & \textbf{60.78} & -- & -- & \textbf{72.16} & -- & -- & 62.10 & -- & -- & 70.10 \\
         MuG \texttt{\small RoBERTa} & 64.18 & 57.03 & 60.33 & 70.47 & 71.88 & 71.16 & 63.78 & 61.88 & 62.79 & 68.61 & 72.20 & 70.34 \\
         \quad w/ \textsc{\small StructApt-Dep} & 64.18 & 56.41 & 60.03 & 71.62 & 71.92 & \underline{71.72} & 63.96 & 61.67 & 62.70 & 68.85 & 71.81 & 70.28 \\
         \quad w/ \textsc{\small StructApt-Rel} & 64.12 & 57.16 & \underline{60.53} & 73.26 & 71.93 & \underline{71.17} & 62.86 & 63.82 & \underline{\textbf{63.12}} & 69.15 & 74.12 & \underline{\textbf{70.44}} \\
         \bottomrule
    \end{tabular}
    }
    \caption{Benchmarking evaluation results. The marker \textsuperscript{*} indicates results of the model are cited from~\citet{Xu20}. The marker \textsuperscript{\dag} indicates results of the model are cited from its original paper. The marker \textsuperscript{\ddag} indicates results of the model are reproduced from its released code. Results are replaced by -- to indicate they are not available. F1 scores are \underline{underlined} to indicate they outperform their adapter-ablated counterparts. F1 scores are \textbf{boldfaced} to indicate they are the best-performing ones in their areas.}
    \label{tab2}
\end{table*}

Following the common practice in the area, we adopt the exact match precision, recall, and F1 scores as the evaluation metrics. Namely, only when the corresponding elements from two triplets exactly match each other, will it be counted as one match. Further, to gain a robust evaluation, we average values over 10 runs and employ the mean value as the final number.

\subsection{Performance Analysis}

From the results presented in Table~\ref{tab2}, we discover two key findings. The first is that the structural adapter incorporated with the relative distance can primarily improve performance across different models and different PLMs, though the improvements over \texttt{RoBERTa} are not as consistent as those over \texttt{BERT} on different datasets. The second is that the previous SOTA models are further boosted by the structural adapter and yield new SOTA results. These findings generally indicate the \textit{effectiveness}, thus \textit{necessity}, of the structural adapter. Conversely, the dependency distance is prone to parsing errors and sometimes underperforms the relative distance.

Moreover, we surprisingly observe that MuG with \texttt{RoBERTa} is a relatively strong baseline even compared with those remarkable generative ASTE models. Concretely, MuG with \texttt{RoBERTa} approximates or outperforms GAS with \texttt{T5} in terms of F1 scores. This phenomenon encourages some retrospectives on whether generative ASTE models are superior to extractive ones, or the superiority is resulted by the generative PLM.

It can be arguable that the improvements of the structural adapter are marginal; however, we conjecture the inherent reason is that the data for evaluation is of small scale. According to the aforementioned unstable behavior of the adapter when encountering small-scale data in Section~\ref{sec3.3}, we think the evaluation is under-confident and therefore conduct a large-scale evaluation in Section~\ref{sec4} to verify the guess and to get more confident results.

\subsection{Parameter Analysis}

We examine the gap between the structural adapter and structural layer (with dependency distance or relative distance, referred to as \textsc{\small StructLyr-Dep} and \textsc{\small StructLyr-Rel} respectively) from the perspective of incremental parameter scale. The structural layer is exactly a stack of additional transformer layers built upon the PLM, each of which is applied with the structural adapter. The best number of stacked layers is 2 in our pilot study.

\begin{table}[t]
    \centering
    \resizebox{0.47\textwidth}{!}{
    \begin{tabular}{lrcc}
         \toprule
         \textbf{Model} & \textbf{\#Params+} & \textbf{L14} & \textbf{R14} \\
         \midrule
         MuG \texttt{\small BERT} & 0.00 M & 55.06 & 68.00 \\
         \quad w/ \textsc{\small StructLyr-Dep} & 14.17 M & 52.52 & 67.03 \\
         \quad w/ \textsc{\small StructLyr-Rel} & 14.17 M & 51.57 & 67.04 \\
         \quad w/ \textsc{\small StructApt-Dep} & 0.01 M & 55.95 & 68.27 \\
         \quad w/ \textsc{\small StructApt-Rel} & 0.01 M & 55.75 & 68.50 \\
         \bottomrule
    \end{tabular}
    }
    \caption{Parameter comparison. F1 scores are reported. \#Params+ is short for number of incremental parameters.}
    \label{tab3}
\end{table}

The incremental parameters in Table~\ref{tab3} mean that additional parameters are brought to MuG.  The structural adapter achieves 1,000$\times$ scale-down without performance loss compared with the structural layer. Contrarily, it seems that structural layers risk the model on the under-fitting issue due to over-parameterization and get degraded performance compared with MuG. We hereby argue that \textit{parameter efficiency} of the structural adapter is permissible. 

\subsection{Latency Analysis}

To better understand the difference between latency consumed by dependency distance derivation and relative distance derivation. We test the latency caused by the above two derivation procedures. 

While dependency distance derivation costs around 4 micro-seconds per token (250 tokens/ms in other words), relative distance derivation only spends 3e-3 micro-seconds per token (333,000 tokens/ms in other words). That is, the relative distance derivation enjoys a 1,000$\times$ speed-up compared with the dependency distance derivation. Hence, the \textit{latency efficiency} of relative distance derivation is numerically verified.

\section{Large-scale Evaluation}
\label{sec4}

\subsection{Data}

Being aware that the above benchmarking evaluation may be under-confident considering that the data is of small scale, we release a large-scale ASTE dataset, short-named \textbb{Lasted}. The data is collected from one of the largest review platform in China, namely DianPing. After necessary pre-processing steps, these reviews are manually annotated by 10 proficient assessors. For sanity, double-check on these annotations is carried out by a researcher who has devoted herself to the area for years.

For clarity, the pre-processing steps include:
1) removing user identities for privacy consideration; 2) chunking the reviews to shorter examples as they are generally too long (e.g., longer than 512); 3) tokenizing these examples; 4) removing examples without annotations, with less than 4 tokens, or with more than 128 tokens; 5) removing triplets in an example if the triplet has more than 8 tokens in the aspect or has more than 16 tokens in the target, for a too long aspect or opinion indicates the triplet may be not well annotated. Ultimately, these examples are formatted in the format we mentioned in Section~\ref{sec2.1}.

We attain the dataset with a total of 27,835 examples. We uniformly split it into train, development, and test sets with a ratio of 7: 1: 2. The statistics are shown in Table~\ref{tab4}, where we also include \textbf{SemEval R14} for comparison purpose. From the statistics, we can summarize that \textbb{Lasted} is a much larger dataset with longer sentences, which sets a more challenging benchmark for models to achieve a high performance.

\begin{table}
    \centering
    \resizebox{0.47\textwidth}{!}{
    \begin{tabular}{cccccc}
         \toprule
         \multicolumn{2}{c}{\textbf{Dataset}} & \textbf{\#S} & \textbf{\#T} & \textbf{\#T/S} & \textbf{\#Tk/S} \\
         \midrule
         \multirow{3}{*}{\textbf{SemEval R14}} & train & 1266 & 2336 & 1.85 & 17.31 \\
         \cmidrule{2-6}
         & dev & 310 & 577 & 1.86 & 15.81 \\
         \cmidrule{2-6}
         & test & 492 & 994 & 2.02 & 16.34 \\
         \midrule
         \multirow{3}{*}{\textbb{Lasted}} & train & 19485 & 38050 & 1.95 & 34.94 \\
         \cmidrule{2-6}
         & dev & 2783 & 5334 & 1.92 & 34.88 \\
         \cmidrule{2-6}
         & test & 5567 & 10820 & 1.94 & 35.04 \\
         \bottomrule
    \end{tabular}
    }
    \caption{Statistics of \textbb{Lasted}, with a comparison with \textbf{SemEval R14}. \#S denotes number of sentences, \#T denotes number of triplets, \#T/S denotes average number of triplets per sentence, and \#Tk/S denotes average number of tokens per sentence.}
    \label{tab4}
\end{table}

\subsection{Models}

We conduct experiments based on GTS and MuG. While we only test GTS with \texttt{BERT-base}, we further test MuG with \texttt{BERT-base}, \texttt{RoBERTa-base}, and tentatively with \texttt{RoBERTa-large}. As we know that only \texttt{BERT-base} is officially released by~\citet{Devlin19} for Chinese, we retrieve \texttt{RoBERTa-base} and \texttt{RoBERTa-large} released by~\citet{Cui19} on Hugging Face.\footnote{Please see \url{https://huggingface.co/hfl/chinese-roberta-wwm-ext} for more information.}

\subsection{Implementation and Metrics}

The implementation and metrics strictly follow those used in the benchmarking evaluation, with exceptions listed in Appendix~\ref{app2}.

\subsection{Analysis}

We can see from Table~\ref{tab5} that the adapter is still promising under large-scale evaluation. With the notice that the evaluation results should be more confident, we hence can safely conclude that the structural adapter is effective and structural bias is \textit{a necessity} for ASTE even in the context of PLMs. However, the metrics on \textbb{Lasted} are consistently lower than expected, implying the deployment of ASTE systems is still challenging. The structural adapter does not improve \texttt{RoBERTa-large}, we leave the question of how to combine it with large PLMs for future work.

\begin{table}[t]
    \centering
    \resizebox{0.45\textwidth}{!}{
    \begin{tabular}{lccc}
         \toprule
         \multirow{2}{*}{\textbf{Model}} & \multicolumn{3}{c}{\textbb{Lasted}} \\
         \cmidrule{2-4}
         & \textbf{P} & \textbf{R} & \textbf{F}\textsubscript{1} \\
         \midrule
         GTS \texttt{\small BERT-base} & 43.81 & 46.11 & 44.92 \\
         \quad w/ \textsc{\small StructApt-Rel} & 45.38 & 46.22 & \underline{45.79} \\
         MuG \texttt{\small BERT-base} & 47.20 & 45.28 & 46.22 \\
         \quad w/ \textsc{\small StructApt-Rel} & 49.64 & 45.02 & \underline{47.22} \\
         MuG \texttt{\small RoBERTa-base} & 48.10 &  44.98 & 46.49 \\ 
         \quad w/ \textsc{\small StructApt-Rel} & 50.40 & 44.77 & \underline{47.42} \\
         MuG \texttt{\small RoBERTa-large} & 49.49 & 46.85 & 48.13 \\
         \quad w/ \textsc{\small StructApt-Rel} & 48.33 & 47.91 & 48.13 \\
         \bottomrule
    \end{tabular}
    }
    \caption{Large-scale evaluation results on \textbb{Lasted}. F1 scores are \underline{underlined} to indicate they outperform their adapter-ablated counterparts.}
    \label{tab5}
\end{table}

\section{Related Work}

\subsection{Aspect Sentiment Triplet Extraction}

Aspect sentiment triplet extraction is a recently proposed task to extract aspects, opinions, and sentiment relations~\citep{Peng20}, serving as a complete solution to aspect sentiment analysis~\citep{ZhangLS19,Ma22}. While the first-ever work delving into the task takes a pipeline system, succeeding work shifts their attention from pipeline models to joint models. \citet{Zhang20} and~\citet{Wu20} share similar spirits to treat three sub-tasks in a multi-task manner. Specifically,~\citet{Wu20} proposes to consider the extraction of three elements in a unified grid tagging scheme. Later studies exploit inductive biases such as span-level interactions~\citep{Xu21} and structural bias~\citep{Chen21b}. To our surprise, none of them inspects whether inductive biases, particularly structural bias, are significant for PLM-enhanced ASTE models. Our work seeks to answer this question by putting forward a flexible adapter and checking whether the adapter is a necessity.

\subsection{Adapter for PLM}

An adapter is an emergent concept which means an efficient module injected into the PLM so that the PLM can better adapt to downstream tasks~\citep{Houlsby19}. Applications including speed translation~\citep{Le21}, language transfer~\citep{He21}, etc. have been witnessed. Traditionally, parameters of the PLM should not be tuned during fine-tuning once the adapter is armed. Nevertheless, recent work~\citep{Liu21} finds that when injecting discrete information, unfreezing the parameters of the PLM will bring further performance gain. While previous adapters are modules and thus far from truly lightweight, we propose to leverage the structured attention as a sort of adapter, which is lightweight. 

\section{Conclusion}

In this paper, we are concerned about the parameter and latency inefficiency issues of incorporating structural bias to PLMs for aspect sentiment triplet extraction, and raise the question on whether structural bias is a necessity. To answer the question, we propose to use an adapter to integrate the relative position structure into PLMs for a light parameter demand compared with incremental layers and low latency compared with the syntactic dependency structure. We carry out benchmarking experiments on SemEval benchmarks and large-scale experiments on our newly released \textbb{Lasted} dataset as a supplementary. The results in two rounds of evaluations show that the structural adapter is an appealing choice regarding its effectiveness, parameter efficiency, and latency efficiency, implying the structural bias, in the form of the structural adapter, is a necessity even with PLMs.

\section*{Acknowledgements}

This research was supported in part by Natural Science Foundation of Beijing (grant number:  4222036) and Huawei Technologies (grant number: TC20201228005).

\bibliography{anthology,custom}

\begin{thebibliography}{33}
\expandafter\ifx\csname natexlab\endcsname\relax\def\natexlab#1{#1}\fi

\bibitem[{Chen et~al.(2021{\natexlab{a}})Chen, Wang, Liu, and Wang}]{Chen21a}
Shaowei Chen, Yu~Wang, Jie Liu, and Yuelin Wang. 2021{\natexlab{a}}.
\newblock \href {https://ojs.aaai.org/index.php/AAAI/article/view/17500}
  {Bidirectional machine reading comprehension for aspect sentiment triplet
  extraction}.
\newblock In \emph{AAAI}, pages 12666--12674.

\bibitem[{Chen et~al.(2021{\natexlab{b}})Chen, Huang, Liu, Shi, and
  Jin}]{Chen21b}
Zhexue Chen, Hong Huang, Bang Liu, Xuanhua Shi, and Hai Jin.
  2021{\natexlab{b}}.
\newblock \href {https://doi.org/10.18653/v1/2021.findings-acl.128} {Semantic
  and syntactic enhanced aspect sentiment triplet extraction}.
\newblock In \emph{ACL}, pages 1474--1483.

\bibitem[{Cui et~al.(2019)Cui, Che, Liu, Qin, Yang, Wang, and Hu}]{Cui19}
Yiming Cui, Wanxiang Che, Ting Liu, Bing Qin, Ziqing Yang, Shijin Wang, and
  Guoping Hu. 2019.
\newblock \href {http://arxiv.org/abs/1906.08101} {Pre-training with whole word
  masking for {C}hinese {BERT}}.
\newblock \emph{arXiv}, abs/1906.08101.

\bibitem[{Dai et~al.(2021)Dai, Yan, Sun, Liu, and Qiu}]{Dai21}
Junqi Dai, Hang Yan, Tianxiang Sun, Pengfei Liu, and Xipeng Qiu. 2021.
\newblock \href {https://doi.org/10.18653/v1/2021.naacl-main.146} {Does syntax
  matter? {A} strong baseline for aspect-based sentiment analysis with
  roberta}.
\newblock In \emph{NAACL}, pages 1816--1829.

\bibitem[{Devlin et~al.(2019)Devlin, Chang, Lee, and Toutanova}]{Devlin19}
Jacob Devlin, Ming{-}Wei Chang, Kenton Lee, and Kristina Toutanova. 2019.
\newblock \href {https://doi.org/10.18653/v1/n19-1423} {{BERT:} pre-training of
  deep bidirectional transformers for language understanding}.
\newblock In \emph{NAACL}, pages 4171--4186.

\bibitem[{Dozat and Manning(2017)}]{Dozat17}
Timothy Dozat and Christopher~D. Manning. 2017.
\newblock \href {https://openreview.net/forum?id=Hk95PK9le} {Deep biaffine
  attention for neural dependency parsing}.
\newblock In \emph{ICLR}.

\bibitem[{He et~al.(2021)He, Liu, Ye, Tan, Ding, Cheng, Low, Bing, and
  Si}]{He21}
Ruidan He, Linlin Liu, Hai Ye, Qingyu Tan, Bosheng Ding, Liying Cheng,
  Jia{-}Wei Low, Lidong Bing, and Luo Si. 2021.
\newblock \href {https://doi.org/10.18653/v1/2021.acl-long.172} {On the
  effectiveness of adapter-based tuning for pretrained language model
  adaptation}.
\newblock In \emph{ACL}, pages 2208--2222.

\bibitem[{Houlsby et~al.(2019)Houlsby, Giurgiu, Jastrzebski, Morrone,
  de~Laroussilhe, Gesmundo, Attariyan, and Gelly}]{Houlsby19}
Neil Houlsby, Andrei Giurgiu, Stanislaw Jastrzebski, Bruna Morrone, Quentin
  de~Laroussilhe, Andrea Gesmundo, Mona Attariyan, and Sylvain Gelly. 2019.
\newblock \href {http://proceedings.mlr.press/v97/houlsby19a.html}
  {Parameter-efficient transfer learning for {NLP}}.
\newblock In \emph{ICML}, pages 2790--2799.

\bibitem[{Huang et~al.(2021)Huang, Wang, Li, Liu, Zhang, Cheng, Yin, and
  Wang}]{Huang21}
Lianzhe Huang, Peiyi Wang, Sujian Li, Tianyu Liu, Xiaodong Zhang, Zhicong
  Cheng, Dawei Yin, and Houfeng Wang. 2021.
\newblock \href {https://arxiv.org/abs/2102.08549} {First target and opinion
  then polarity: Enhancing target-opinion correlation for aspect sentiment
  triplet extraction}.
\newblock \emph{arXiv}, abs/2102.08549.

\bibitem[{Le et~al.(2021)Le, Pino, Wang, Gu, Schwab, and Besacier}]{Le21}
Hang Le, Juan~Miguel Pino, Changhan Wang, Jiatao Gu, Didier Schwab, and Laurent
  Besacier. 2021.
\newblock \href {https://doi.org/10.18653/v1/2021.acl-short.103} {Lightweight
  adapter tuning for multilingual speech translation}.
\newblock In \emph{ACL}, pages 817--824.

\bibitem[{Lewis et~al.(2020)Lewis, Liu, Goyal, Ghazvininejad, Mohamed, Levy,
  Stoyanov, and Zettlemoyer}]{Lewis20}
Mike Lewis, Yinhan Liu, Naman Goyal, Marjan Ghazvininejad, Abdelrahman Mohamed,
  Omer Levy, Veselin Stoyanov, and Luke Zettlemoyer. 2020.
\newblock \href {https://doi.org/10.18653/v1/2020.acl-main.703} {{BART:}
  denoising sequence-to-sequence pre-training for natural language generation,
  translation, and comprehension}.
\newblock In \emph{ACL}, pages 7871--7880.

\bibitem[{Liu et~al.(2021)Liu, Fu, Zhang, and Xiao}]{Liu21}
Wei Liu, Xiyan Fu, Yue Zhang, and Wenming Xiao. 2021.
\newblock \href {https://doi.org/10.18653/v1/2021.acl-long.454} {Lexicon
  enhanced {C}hinese sequence labeling using {BERT} adapter}.
\newblock In \emph{ACL}, pages 5847--5858.

\bibitem[{Liu et~al.(2019)Liu, Ott, Goyal, Du, Joshi, Chen, Levy, Lewis,
  Zettlemoyer, and Stoyanov}]{Liu19}
Yinhan Liu, Myle Ott, Naman Goyal, Jingfei Du, Mandar Joshi, Danqi Chen, Omer
  Levy, Mike Lewis, Luke Zettlemoyer, and Veselin Stoyanov. 2019.
\newblock \href {http://arxiv.org/abs/1907.11692} {Roberta: {A} robustly
  optimized {BERT} pretraining approach}.
\newblock \emph{arXiv}, abs/1907.11692.

\bibitem[{Ma et~al.(2021)Ma, Zhang, and Song}]{Ma21}
Fang Ma, Chen Zhang, and Dawei Song. 2021.
\newblock \href {https://doi.org/10.18653/v1/2021.findings-acl.116} {Exploiting
  position bias for robust aspect sentiment classification}.
\newblock In \emph{ACL}, pages 1352--1358.

\bibitem[{Ma et~al.(2022)Ma, Zhang, Zhang, and Song}]{Ma22}
Fang Ma, Chen Zhang, Bo~Zhang, and Dawei Song. 2022.
\newblock \href {https://doi.org/10.48550/arXiv.2207.08099} {Aspect-specific
  context modeling for aspect-based sentiment analysis}.
\newblock \emph{arXiv}, abs/2207.08099.

\bibitem[{Peng et~al.(2020)Peng, Xu, Bing, Huang, Lu, and Si}]{Peng20}
Haiyun Peng, Lu~Xu, Lidong Bing, Fei Huang, Wei Lu, and Luo Si. 2020.
\newblock \href {https://aaai.org/ojs/index.php/AAAI/article/view/6383}
  {Knowing what, how and why: {A} near complete solution for aspect-based
  sentiment analysis}.
\newblock In \emph{AAAI}, pages 8600--8607.

\bibitem[{Pontiki et~al.(2014)Pontiki, Galanis, Pavlopoulos, Papageorgiou,
  Androutsopoulos, and Manandhar}]{Pontiki14}
Maria Pontiki, Dimitris Galanis, John Pavlopoulos, Harris Papageorgiou, Ion
  Androutsopoulos, and Suresh Manandhar. 2014.
\newblock \href {https://doi.org/10.3115/v1/s14-2004} {Semeval-2014 task 4:
  Aspect based sentiment analysis}.
\newblock In \emph{SemEval}, pages 27--35.

\bibitem[{Qi et~al.(2020)Qi, Zhang, Zhang, Bolton, and Manning}]{Qi20}
Peng Qi, Yuhao Zhang, Yuhui Zhang, Jason Bolton, and Christopher~D. Manning.
  2020.
\newblock \href {https://doi.org/10.18653/v1/2020.acl-demos.14} {Stanza: {A}
  python natural language processing toolkit for many human languages}.
\newblock In \emph{ACL}, pages 101--108.

\bibitem[{Raffel et~al.(2020)Raffel, Shazeer, Roberts, Lee, Narang, Matena,
  Zhou, Li, and Liu}]{Raffel20}
Colin Raffel, Noam Shazeer, Adam Roberts, Katherine Lee, Sharan Narang, Michael
  Matena, Yanqi Zhou, Wei Li, and Peter~J. Liu. 2020.
\newblock \href {http://jmlr.org/papers/v21/20-074.html} {Exploring the limits
  of transfer learning with a unified text-to-text transformer}.
\newblock \emph{JMLR}, 21:140:1--140:67.

\bibitem[{Shaw et~al.(2018)Shaw, Uszkoreit, and Vaswani}]{Shaw18}
Peter Shaw, Jakob Uszkoreit, and Ashish Vaswani. 2018.
\newblock \href {https://doi.org/10.18653/v1/n18-2074} {Self-attention with
  relative position representations}.
\newblock In \emph{NAACL}, pages 464--468.

\bibitem[{Sun et~al.(2017)Sun, Wu, Lan, Sun, and Zhang}]{Sun17}
Changzhi Sun, Yuanbin Wu, Man Lan, Shiliang Sun, and Qi~Zhang. 2017.
\newblock \href {https://doi.org/10.18653/v1/e17-1097} {Large-scale opinion
  relation extraction with distantly supervised neural network}.
\newblock In \emph{EACL}, pages 1033--1043.

\bibitem[{Vaswani et~al.(2017)Vaswani, Shazeer, Parmar, Uszkoreit, Jones,
  Gomez, Kaiser, and Polosukhin}]{Vaswani17}
Ashish Vaswani, Noam Shazeer, Niki Parmar, Jakob Uszkoreit, Llion Jones,
  Aidan~N. Gomez, Lukasz Kaiser, and Illia Polosukhin. 2017.
\newblock \href
  {https://proceedings.neurips.cc/paper/2017/hash/3f5ee243547dee91fbd053c1c4a845aa-Abstract.html}
  {Attention is all you need}.
\newblock In \emph{NeurIPS}, pages 5998--6008.

\bibitem[{Wang et~al.(2020)Wang, Shin, Liu, Polozov, and Richardson}]{Wang20}
Bailin Wang, Richard Shin, Xiaodong Liu, Oleksandr Polozov, and Matthew
  Richardson. 2020.
\newblock \href {https://doi.org/10.18653/v1/2020.acl-main.677} {{RAT-SQL:}
  relation-aware schema encoding and linking for text-to-sql parsers}.
\newblock In \emph{ACL}, pages 7567--7578.

\bibitem[{Wu et~al.(2009)Wu, Zhang, Huang, and Wu}]{Wu09}
Yuanbin Wu, Qi~Zhang, Xuanjing Huang, and Lide Wu. 2009.
\newblock \href {https://aclanthology.org/D09-1159/} {Phrase dependency parsing
  for opinion mining}.
\newblock In \emph{EMNLP}, pages 1533--1541.

\bibitem[{Wu et~al.(2020{\natexlab{a}})Wu, Ying, Zhao, Fan, Dai, and
  Xia}]{Wu20}
Zhen Wu, Chengcan Ying, Fei Zhao, Zhifang Fan, Xinyu Dai, and Rui Xia.
  2020{\natexlab{a}}.
\newblock \href {https://doi.org/10.18653/v1/2020.findings-emnlp.234} {Grid
  tagging scheme for end-to-end fine-grained opinion extraction}.
\newblock In \emph{EMNLP}, pages 2576--2585.

\bibitem[{Wu et~al.(2020{\natexlab{b}})Wu, Chen, Kao, and Liu}]{Wu20b}
Zhiyong Wu, Yun Chen, Ben Kao, and Qun Liu. 2020{\natexlab{b}}.
\newblock \href {https://doi.org/10.18653/v1/2020.acl-main.383} {Perturbed
  masking: Parameter-free probing for analyzing and interpreting {BERT}}.
\newblock In \emph{ACL}, pages 4166--4176.

\bibitem[{Xu et~al.(2021)Xu, Chia, and Bing}]{Xu21}
Lu~Xu, Yew~Ken Chia, and Lidong Bing. 2021.
\newblock \href {https://doi.org/10.18653/v1/2021.acl-long.367} {Learning
  span-level interactions for aspect sentiment triplet extraction}.
\newblock In \emph{ACL}, pages 4755--4766.

\bibitem[{Xu et~al.(2020)Xu, Li, Lu, and Bing}]{Xu20}
Lu~Xu, Hao Li, Wei Lu, and Lidong Bing. 2020.
\newblock \href {https://doi.org/10.18653/v1/2020.emnlp-main.183}
  {Position-aware tagging for aspect sentiment triplet extraction}.
\newblock In \emph{EMNLP}, pages 2339--2349.

\bibitem[{Yan et~al.(2021)Yan, Dai, Ji, Qiu, and Zhang}]{Yan21}
Hang Yan, Junqi Dai, Tuo Ji, Xipeng Qiu, and Zheng Zhang. 2021.
\newblock \href {https://doi.org/10.18653/v1/2021.acl-long.188} {A unified
  generative framework for aspect-based sentiment analysis}.
\newblock In \emph{ACL}, pages 2416--2429.

\bibitem[{Zhang et~al.(2019{\natexlab{a}})Zhang, Li, and Song}]{ZhangLS19}
Chen Zhang, Qiuchi Li, and Dawei Song. 2019{\natexlab{a}}.
\newblock \href {https://doi.org/10.18653/v1/D19-1464} {Aspect-based sentiment
  classification with aspect-specific graph convolutional networks}.
\newblock In \emph{EMNLP}, pages 4567--4577.

\bibitem[{Zhang et~al.(2019{\natexlab{b}})Zhang, Li, and Song}]{Zhang19}
Chen Zhang, Qiuchi Li, and Dawei Song. 2019{\natexlab{b}}.
\newblock \href {https://doi.org/10.1145/3331184.3331351} {Syntax-aware
  aspect-level sentiment classification with proximity-weighted convolution
  network}.
\newblock In \emph{SIGIR}, pages 1145--1148.

\bibitem[{Zhang et~al.(2020)Zhang, Li, Song, and Wang}]{Zhang20}
Chen Zhang, Qiuchi Li, Dawei Song, and Benyou Wang. 2020.
\newblock \href {https://doi.org/10.18653/v1/2020.findings-emnlp.72} {A
  multi-task learning framework for opinion triplet extraction}.
\newblock In \emph{EMNLP}, pages 819--828.

\bibitem[{Zhang et~al.(2021)Zhang, Li, Deng, Bing, and Lam}]{Zhang21}
Wenxuan Zhang, Xin Li, Yang Deng, Lidong Bing, and Wai Lam. 2021.
\newblock \href {https://doi.org/10.18653/v1/2021.acl-short.64} {Towards
  generative aspect-based sentiment analysis}.
\newblock In \emph{ACL}, pages 504--510.

\end{thebibliography}
\bibliographystyle{acl_natbib}


\appendix

\section{Data Statistics of SemEval}
\label{app1}

\begin{table}[ht]
    \centering
    \resizebox{0.40\textwidth}{!}{
    \begin{tabular}{cccccc}
         \toprule
         \multicolumn{2}{c}{\textbf{Dataset}} & \textbf{\#S} & \textbf{\#T} & \textbf{\#T/S} & \textbf{\#Tk/S} \\
         \midrule
         \multirow{3}{*}{\textbf{L14}} & train & 906 & 1460 & 1.61 & 19.15 \\
         \cmidrule{2-6}
         & dev & 219 & 346 & 1.58 & 19.06 \\
         \cmidrule{2-6}
         & test & 328 & 543 & 1.66 & 15.77 \\
         \midrule
         \multirow{3}{*}{\textbf{R14}} & train & 1266 & 2336 & 1.85 & 17.31 \\
         \cmidrule{2-6}
         & dev & 310 & 577 & 1.86 & 15.81 \\
         \cmidrule{2-6}
         & test & 492 & 994 & 2.02 & 16.34 \\
         \midrule
         \multirow{3}{*}{\textbf{R15}} & train & 605 & 1013 & 1.67 & 14.80 \\
         \cmidrule{2-6}
         & dev & 148 & 249 & 1.68 & 14.34 \\
         \cmidrule{2-6}
         & test & 322 & 485 & 1.51 & 15.63 \\
         \midrule
         \multirow{3}{*}{\textbf{R16}} & train & 857 & 1394 & 1.63 & 15.15 \\
         \cmidrule{2-6}
         & dev & 210 & 339 & 1.61 & 14.16 \\
         \cmidrule{2-6}
         & test & 326 & 514 & 1.58 & 14.70 \\
         \bottomrule
    \end{tabular}
    }
    \caption{Statistics of four datasets from \textbf{SemEval}. \#S denotes number of sentences, \#T denotes number of triplets, \#T/S denotes average number of triplets per sentence, and \#Tk/S denotes average number of tokens per sentence. L denotes laptop domain while R denotes restaurant domain.}
    \label{tab1}
\end{table}

\section{Full Implementation Details}
\label{app2}

Our models are implemented with PyTorch and verified on an Nvidia V100, and they are generally trained with following instructions. 

For parameter settings in the benchmarking evaluation, the batch size is 8 for models without the adapter, whereas it is 6 for models with the adapter for stability, and the maximum norm for gradients is 1. The learning rate is set hierarchically, where the learning rate for the PLM and adapter is searched with \{1e-5,2e-5,3e-5,5e-5\} while that for the triplet parser is set 10 times of the former. The training procedure is scheduled as such: the number of maximum training epochs is 20, and the number of patience epochs is 5. Learning rates are warmed up for the first 2 epochs and decayed for the rest epochs. The threshold for the maximum distance $\tau$ is 8.

For parameter settings in the large-scale evaluation, the batch size is accordingly doubled, since we have data of a much larger scale. 

\end{document}